\documentclass[letterpaper, 10 pt, conference]{ieeeconf}  % Comment this line out
\IEEEoverridecommandlockouts                              % This command is only
\usepackage[utf8]{inputenc}
\usepackage[T1]{fontenc}
\usepackage{graphicx}
\usepackage{amsmath}
\usepackage{listings}

\usepackage[dvipsnames,svgnames,x11names]{xcolor}
\usepackage[final]{changes}
\title{\LARGE \bf
Learning-Augmented Model-Based Planning for Visual Exploration}

\author{{Yimeng Li*$^1$}, {Arnab Debnath*$^1$}, {Gregory J. Stein$^1$} and {Jana Ko{\v{s}}eck{\'a}$^1$}
\thanks{* Denotes equal contribution.}
\thanks{$^{1}$Yimeng Li, Arnab Debnath, Gregory Stein and Jana Kosecka are with the Department of Computer Science, 
George Mason University, 
4400 University Dr, Fairfax, VA, USA {\tt\small [yli44, adebnath, gjstein, kosecka]@gmu.edu}}}

\begin{document}

\maketitle
\thispagestyle{empty}
\pagestyle{empty}

\begin{abstract}
We consider the problem of time-limited robotic exploration in previously unseen environments where exploration is limited by a predefined amount of time. 
We propose a novel exploration approach using learning-augmented model-based planning. 
We generate a set of subgoals associated with frontiers on the current map and derive a Bellman Equation for exploration with these subgoals.
Visual sensing and advances in semantic mapping of indoor scenes are exploited for training a deep convolutional neural network to estimate properties associated with each frontier: the expected unobserved area beyond the frontier and the expected timesteps (discretized actions) required to explore it.
The proposed model-based planner is guaranteed to explore the whole scene if time permits.
We thoroughly evaluate our approach on a large-scale pseudo-realistic indoor dataset (Matterport3D) with the Habitat simulator. 
We compare our approach with classical and more recent RL-based exploration methods. 
Our approach surpasses the greedy strategies by 2.1$\%$ and the RL-based exploration methods by 8.4$\%$ in terms of coverage.
\end{abstract}

\section{Introduction}
We consider a robot deployed in a previously-unseen environment and given a limited time during which it is expected to explore, move around, and become familiar with its surroundings.
Task-independent exploration is critical for many downstream navigation tasks, including traveling to a point goal~\cite{Ramakrishnan2020AnEO, Chaplot2020NeuralTS, Georgakis2022UncertaintydrivenPF}, looking for a target object~\cite{ramakrishnan2022poni}, or disaster response scenarios~\cite{bradley2021learning}.
The robot must explore intelligently to cover the maximum area under time constraints.

However, efficient exploration is incredibly challenging since good performance requires inferences about unseen space's layout.
For example, a few steps of travel inside a living room can bring a broad vision of the entire scene rather than spending a long time walking through a narrow corridor.
Due to the complexities of planning under uncertainty, many existing approaches rely on learning to make predictions about unseen space and determine behavior.
Recent advent of learning-based approaches proposed for robot visual exploration~\cite{Ramakrishnan2020AnEO,   Georgakis2022UncertaintydrivenPF, chaplot2020learning, chen2018learning} leverages 3D textured mapped models of indoor scenes collected from real-world environments~\cite{Chang2017Matterport3DLF, xia2018gibson} to learn structural and semantic regularities informative for exploration.

Many learning-driven strategies for exploration~\cite{ Ramakrishnan2020AnEO, chen2018learning} are model-free and trained via deep reinforcement learning.
These models typically learn policies on top of intermediate representations of indoor scenes, are data-intensive to train, and require millions of frames of training data.
They often struggle in novel environments as the length of exploration grows.
These approaches are either unreliable and fail to fully explore the environment in the fullness of time or have poor performance, owing to their inability to directly predict the long-horizon impact of their actions.

\begin{figure}[t!]
\centering
\includegraphics[width=0.48\textwidth]{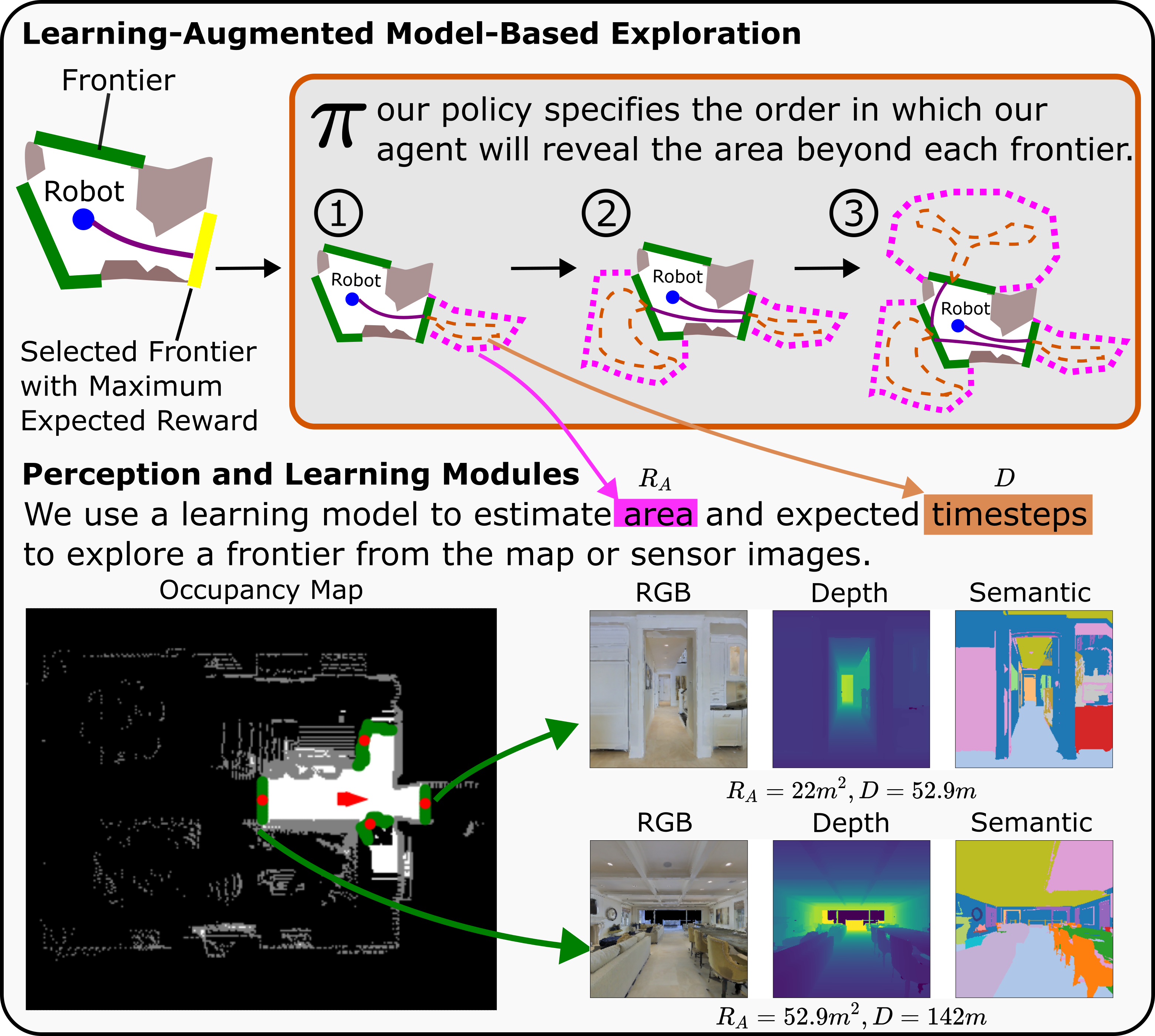}
\vspace{-1.5em}
\caption{
\textbf{Schematic showing our learning-augmented model-based exploration strategy.} 
Our approach associates high-level exploratory actions with \emph{frontiers} and uses a Bellman Equation to estimate the expected reward of each action. 
The learning module takes in the input of the currently observed occupancy map and semantic map or egocentric observations and predicts the unknown properties of the frontier. 
The computed frontiers are illustrated with green pixels in the occupancy map, and the red nodes are centroids.
$R_A$ and $D$ are the estimated area and timesteps for each frontier.
}\vspace{-1em}
\label{fig:title}
\end{figure}

In \emph{frontier-based methods}, the robot's actions correspond to navigation to frontiers---each a boundary between free and unknown space---and rely on the partial map to plan and reliably navigate through space the robot has already seen.
Fig.~\ref{fig:title} (bottom) shows computed frontiers on an example occupancy map.
Classical frontier-based exploration (FBE) methods~\cite{yamauchi1997frontier} rely on an occupancy map, built via a laser range finder, to compute frontiers, and greedily select the next frontier-action via simple heuristics (e.g., navigation to the nearest frontier), resulting in suboptimal plans.
Recent work~\cite{ramakrishnan2022poni} uses visual sensing and replaces simple heuristics with a learning-based module that estimates the expected area beyond each frontier from a top-down-view semantic map. 
Though the predictions from learning help to improve exploration, planning is still myopic, limiting performance.
If such frontier-based planning strategies are to improve, the robot must have the capacity to reason into the future and understand the impact of its actions: including both (i) \emph{how much area exploration beyond each frontier is expected to reveal} and (ii) \emph{how long it will take to explore each region of the environment}.

We present \emph{Learning-Augmented Model-Based Frontier-Based Exploration} (LFE) that allows for reliable long-horizon exploration.
Taking inspiration from recent work by Stein et al.~\cite{stein2018learning}, we introduce a high-level abstraction that allows for model-based planning over the set of available frontiers and approximates a complex Partially Observable Markov Decision Process (POMDP)~\cite{kaelbling1998planning} as a much simpler Markov Decision Process (MDP)~\cite{russell2010artificial}.
Motivated by Ramakrishnan et al.~\cite{ramakrishnan2022poni}, we use environment semantics and visual perception to learn how to estimate the expected area that exploration beyond a frontier will reveal and how long such an exploration is expected to take.
Our resulting model-based planner produces the sequence of exploratory actions that will maximize expected area coverage in the allowable time.
Moreover, our approach generalizes well to large environments, while model-free RL approaches~\cite{Ramakrishnan2020AnEO, chen2018learning} do not accommodate well to large-scale environments not encountered during training.

Our contribution is a frontier-based model, similar to that of Stein et al.~\cite{stein2018learning} for the task of time-limited exploration, with the aim to maximize the explored area within a predefined period. 
We evaluate the performance of our approach on a large-scale pseudo-realistic indoor dataset (Matterport3D~\cite{Chang2017Matterport3DLF}) with the Habitat simulator~\cite{savva2019habitat}, and achieve state-of-the-art performance (80.7\%) on time-limited area coverage and compare our approach with learned greedy~\cite{ramakrishnan2022poni} and RL-based~\cite{chaplot2020learning} methods.

%=============================================================================================================================================
\section{Related Work}
\noindent \textbf{Embodied Agent Navigation}
Embodied agent navigation~\cite{deitke2022retrospectives} refers to an agent's ability to navigate and interact with its environment in a way similar to how a human would. 
In recent years, there has been a surge of research in embodied navigation thanks to the availability of photo-realistic simulators~\cite{xia2018gibson, szot2021habitat, habitat19iccv, kolve2017ai2} for large-scale indoor environments~\cite{Chang2017Matterport3DLF, ramakrishnan2021hm3d, yadav2022habitat}.
The proposed tasks include PointGoal Navigation~\cite{wijmans2019dd, Partsey2022IsMN, li2022comparison}, ObjectGoal Navigation~\cite{pal2021learning, chaplot2020object}, ImageGoal Navigation~\cite{Chaplot2020NeuralTS, savinov2018semi, Hahn2021NoRN, li2020learning} Vision-and-Language Navigation (VLN)~\cite{Krantz2021WaypointMF, krantz2022iterative}, and Object Rearrangement~\cite{batra2020rearrangement, Sarch2022TIDEETU, gu2022multi}.
In this paper, we tackle the exploration task, which is the fundamental problem to all the aforementioned navigation tasks.

\noindent \textbf{End-to-End Visual Exploration}
Recent deep reinforcement learning methods~\cite{Ramakrishnan2020AnEO, chen2018learning, chaplot2020object, fang2019scene} 
learn a policy to predict low-level actions directly from raw RGBD observations or local occupancy maps to improve navigation performance or exploration coverage. 
These end-to-end RL approaches usually are sample inefficient and require high computation.
Instead of directly selecting short-term actions, we introduce a dynamic action set of actions to navigate to a frontier that allow us to make predictions computationally efficiently.

\noindent\textbf{Intermediate Goals}
Other deep reinforcement learning methods~\cite{Georgakis2022UncertaintydrivenPF, chaplot2020learning, ramakrishnan2020occupancy,  chen2020learning} use hierarchical models that have global policies that predict long-term goals and use local navigation modules to reach the long-term goal.
However, such methods still choose long-term goals greedily, and these long-term goals are usually located in an unknown area which may not be reachable by the agent.
Our approach is most similar to that of Stein et al.~\cite{stein2018learning}, in which a learning-augmented frontier-based abstraction is used for point-goal navigation. 
We instead derive a Bellman Equation particularly for the time-limited exploration task.

\noindent\textbf{Frontier-Based Exploration}
Frontier-based exploration is first proposed by Yamauchi~\cite{yamauchi1997frontier}, where the robot simply goes to the nearest frontier. 
With the availability of camera sensors and semantic-informative simulators, it has become possible to estimate more complex properties of frontiers.
Recently, Ramakrishnan et al.~\cite{ramakrishnan2022poni} used visual sensing and proposed a learning-augmented FBE model for the robot going to a target object.
Their learning module is a neural network that takes the observed semantic map as input and estimates the unknown space beyond the frontier and semantic heuristics may lead to the object. 
Our method differs from theirs in that we use a variant of the Bellman Equation to guide frontier selection that considers travel cost besides area of the unknown space.

\begin{figure*}[ht!]
\centering
\includegraphics[width=0.9\textwidth]{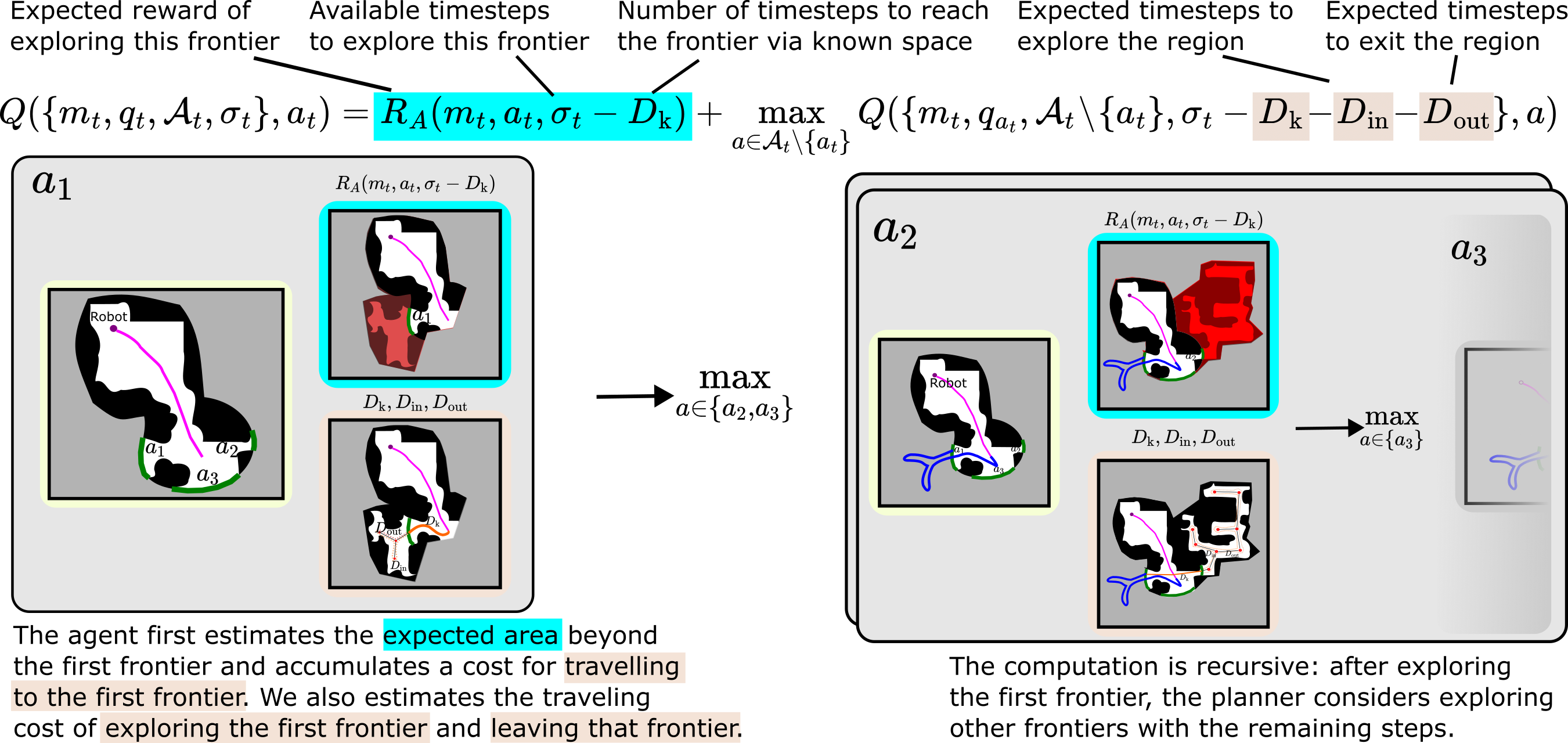}
\vspace{-.5em}
\caption{
\textbf{This diagram gives an overview of our learning-augmented model-based exploration algorithm}, which allows us to compute the expected value of each action in a computationally feasible way for planning through an unknown environment.
We use learning modules to estimate the terms $R_A$, $D_\text{in}$ and $D_\text{out}$, thereby introducing prior knowledge about environment regularities into the decision-making procedure.
}
\label{fig:approach}
\vspace{-.5em}
\end{figure*}

%=============================================================================================================================================
\section{Approach: Learning-Augmented Model-Based Frontier Exploration}
\label{sec:approach}
We aim to maximize area coverage during exploration through a static, unknown environment within an allowed number of timesteps.
We model exploration as a Partially Observable Markov Decision Process (POMDP)~\cite{kaelbling1998planning}.
The expected reward of taking an action can be expressed via a belief-space variant of the Bellman Equation~\cite{Pineau-2002-8519}:
\begin{equation}
\label{eq:pomdp}
\begin{split}
Q(b_t, a_t \in \mathcal{A}(b_t)) & = \sum_{b_{t+1}}P(b_{t+1}|b_t,a_t)[R(b_{t+1}, b_t,a_t) \\
& + \max_{a_{t+1}\in \mathcal{A}(b_{t+1})} Q(b_{t+1}, a_{t+1})]
\end{split}
\end{equation}
where $R(b_{t+1}, b_{t}, a_{t})$ is the expected reward (in units of area coverage) of reaching belief state $b_{t+1}$ from $b_{t}$ by taking action $a_{t}$.
Our belief state $b_t$ is represented as a three-element tuple consisting of the partially observed map $m_t$, the robot pose $q_t$ and the remaining available timesteps  $\sigma_t$: $b_t = \{m_t, q_t, \sigma_t\}$.
In practice, computing the expected reward via Eq.~\eqref{eq:pomdp} is computationally infeasible, since it requires taking an expectation over the distribution of all possible environments and reasoning hundreds of timesteps into the future.

\subsection{Model-Based Planning with Subgoals}

To simplify the calculation of the expected reward, we introduce an abstraction, inspired by that of Stein et al.~\cite{stein2018learning}, in which the robot's actions are long-term behaviors: each associated with exploring a particular region of unseen space.
Under this abstraction, each frontier, a boundary between free and unseen space, is assigned a \emph{subgoal}. A high-level action $a_t \in \mathcal{A}(b_t)$ consists of first navigating through known space to the subgoal of interest---traveling a \emph{known-space distance} $D_\text{k}(m_t, a_t)$---and then revealing the region beyond.
After each such action, the robot will have (i) accumulated a reward corresponding to the amount of area revealed and (ii) expended time to explore, thus decreasing $\sigma_t$ by how long the exploration took.
As the robot reveals more of the environment, it updates its partial map and recomputes the set of available subgoal-actions and replans.

\subsection{Approximating Expected Reward}\label{sec:theory:approximating}
To select the next high-level action $a_t$, we must be able to evaluate it's expected reward.
In the context of exploration, the expected instantaneous reward for a particular high-level exploratory action $a_t$ is the unexplored navigable area beyond that action's corresponding frontier.
This reward ($R_A$) depends on the current partial map $m_t$ and the area that can be explored in the steps remaining ($\sigma_t - D_\text{k}$), as specified by action $a_t$:
\begin{equation}
\label{eq:approx_1}
R_A(m_t, a_t, \sigma_t-D_\text{k}) \equiv \sum_{b_{t+1}}P(b_{t+1}|b_t,a_t) R(b_{t+1}, b_t,a_t) 
\end{equation}
Since we neither have direct access to the belief nor the computational resources to evaluate Eq.~\eqref{eq:approx_1} directly, we will instead estimate $R_A$ using learning described in Sec.~\ref{sec:learning}.

Evaluating the recursive term in Eq.~\eqref{eq:pomdp} remains challenging since updating the map $m_t$ (and therefore computing $b_{t+1}$) remains computationally infeasible.
We instead make a simplifying assumption: we do not directly update the \deleted{map} \added{occupancy grid} during high-level planning and instead prevent the robot from exploring the same space again by keeping track of what has already been explored.
As such, we define the set of future actions $\mathcal{A}_{t+1}$ as the current set of actions minus the one that was just chosen (since we disallow exploration of the same region more than once), so that $\mathcal{A}_{t+1} = \mathcal{A}_{t} \setminus \{ a_t \}$. 
\added{
Therefore, the expected reward of taking action, $a_t$ using the current map $m_t$, would mean getting the instantaneous reward $R_A$ from exploring the associated frontier and then considering the $A_{t+1}$ actions without introducing new frontiers by updating the map.
}

Moreover, as mentioned in Sec.~\ref{sec:theory:approximating}, during an exploratory action $a_t$ the robot first travels a  distance $D_\text{k}$ through known space and then spends a time to explore the unknown space beyond a frontier.
We use the A$^{\!*}$ algorithm to plan the trajectory through known space and compute $D_\text{k}$.
The expected number of steps to explore can be written as the sum of two terms: $D_\text{in}$, the expected number of steps to explore the region, and $D_\text{out}$, the expected number of steps to exit the newly-explored region.
\added{
Splitting the number of steps into two parts enables easier computation of frontier Q-values, as we only need to consider $D_\text{in}$ when we have only one frontier left.
}
Similar to $R_A$ from Eq.~\eqref{eq:approx_1}, we cannot compute $D_\text{in}$ and $D_\text{out}$ exactly as it is computationally expensive and so estimate them from learning.
Furthermore, upon completing its exploration of a region, the robot has moved to the frontier and so its pose change from $q_t$ to $q_{a_t}$.

With these, we can approximate the recursive term from Eq.~\eqref{eq:pomdp} as follows:
\begin{equation}
\label{eq:approx_2}
\begin{split}
\sum_{b_{t+1}} & P(b_{t+1}|b_t,a_t) \max_{a_{t+1}\in \mathcal{A}(b_{t+1})} Q(b_{t+1}, a_{t+1}) \approx \\
& \max_{a\in \mathcal{A}_t \setminus \{a_t\}} Q(\left\{m_t, q_{a_t},  \mathcal{A}_t \! \setminus \! \{a_t\}, \sigma_t-  D_\text{k} \! - \! D_\text{in} \! - \! D_\text{out} \right\}, a)
\end{split}
\end{equation}
Thus, from Eq.~\eqref{eq:pomdp}, \eqref{eq:approx_1} \& \eqref{eq:approx_2} we can rewrite expected reward:
\begin{equation}
\label{eq:dp}
\begin{split}
Q(&\left\{m_t, q_t, \mathcal{A}_t, \sigma_t  \right\},a_t) = R_A(m_t, a_t, \sigma_t-D_\text{k}) + \\
& \max_{a\in \mathcal{A}_t \setminus \{a_t\}} Q(\left\{m_t, q_{a_t},  \mathcal{A}_t \! \setminus \! \{a_t\}, \sigma_t-  D_\text{k} \! - \! D_\text{in} \! - \! D_\text{out} \right\}, a)
\end{split}
\end{equation}

An illustration of how expected cost is computed under our model can be found in Fig.~\ref{fig:approach}.

\begin{figure*}[ht!]
\centering
\includegraphics[width=0.98\textwidth]{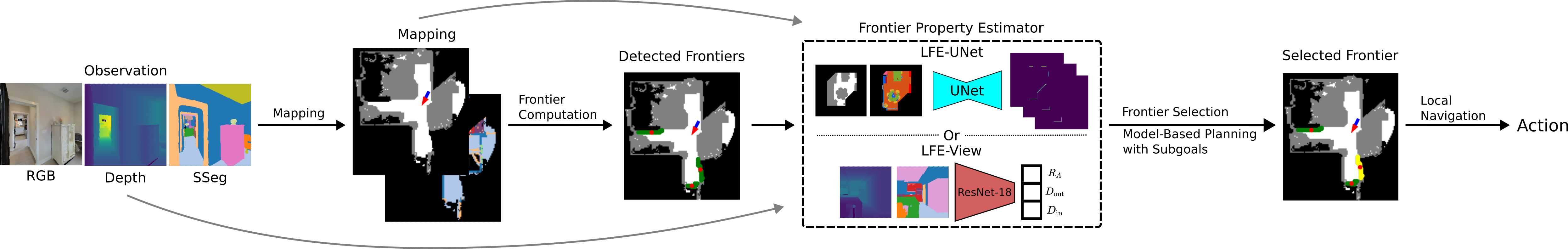}
\vspace{-0.8em}
\caption{
\textbf{Architecture of the entire exploration system.} 
The detected and the selected frontiers are drawn in green and yellow respectively.
Our proposed Learning-augmented model-based frontier-based exploration is used for the frontier selection module.
We design two learning modules to estimate the frontier properties.
The UNet model takes in the local occupancy map and local semantic map and estimates values for all the visible frontiers on the map.
The ResNet-18 model takes in the egocentric depth and semantic observation corresponding to a frontier and estimates the values only for this frontier.
}
\label{fig:modules}
\vspace{-0.5em}
\end{figure*}
\subsection{Hierarchial Navigational Model}
Our planner is hierarchical: high-level planning involves selection among frontier subgoals, while low-level planning involves navigation to the chosen frontier.
Planning with our model proceeds as follows:
(i) we compute the set of frontiers and the respective action set $\mathcal{A}(b_t)$ given the map $m_t$,
(ii) we compute the expected reward for each exploratory action according to Eq.~\eqref{eq:dp},
(iii) we choose the action $a_t^*$ that maximizes the expected reward or, if more than one action reaches the maximum reward, we choose the frontier with the most timesteps remaining.
Once the frontier is selected, 
we (iv) compute a motion plan to travel to the centroid of the frontier associated with action $a_t^*$ ($q^*$) and finally
(v) select the primitive action that best moves along the computed path and update the map given the newly observed area.
This process repeats at each timestep until no frontiers remain in the observed map or time is exhausted.

An overview of our approach can be found in Fig.~\ref{fig:modules}.

%=============================================================================================================================================

\section{Map and Frontier Computation}
\noindent\textbf{Occupancy and Semantic Mapping}\quad{}
The occupancy map is a metric map of size $m \times m$ where each cell of the map can have three values: 0 (unknown), 1 (occupied) and 2 (free space).
The semantic map is a metric map of size $m \times m$ where each cell's value is in the range $[0, N]$.
Each value from 1 to $N$ corresponds to one of the N object categories, and 0 refers to the undetected class. 
Each cell is a $5cm \times 5cm$ region in the real world.
Given an RGB-D view, we build the occupancy map and semantic map by projecting semantic segmentation images to a 3D point cloud using the available depth maps and the robot poses, discretizing the point cloud into a voxel grid and taking the top-down view of the voxel grid.
The occupancy map depends on the height of the points projected to each cell. 
The semantic map depends on the majority category of the points located at the top grid of each cell.

\noindent\textbf{Frontier Computation}\quad{}
Frontiers consist of unknown grid cells on the boundary of free space in the partial occupancy map.
A frontier is a set of connected cells that follow an eight-neighbour connection.
The center of the frontier is the centroid of the points belonging to that frontier.
We further reduce the number of frontiers by filtering out frontiers unreachable from the robot on the current occupancy map.

\noindent\textbf{Local Navigation}\quad{}
The local navigation module uses the Fast Marching Method~\cite{sethian1996fast} to plan a path to the selected frontier from the current location on the occupancy map. 
It takes deterministic actions (move forward by 25cm, turn left/right by 30 degrees) to reach the center of the frontier. 
We update the map after each step and replan. 

\noindent\textbf{Computing Expected Cost}\quad{}
Eq.~\eqref{eq:dp} is evaluated recursively.
To reduce the computation complexity, we select the six frontiers with the most significant area $R_A$ to define the set of possible actions $\mathcal{A}_t$.
Since we need $D_\text{in}$ to reveal the whole area beyond a frontier, if we don't have enough timesteps left ($\sigma = \sigma_t-D_\text{k}$), we only get a portion of the full reward and our reward structure looks like the following:
\begin{equation}
\label{eq:RA}
%R_A(m_t, a_t, \sigma_t-D_\text{k}) = R_A(m_t, a_t) \cdot \min\left(1, \frac{(\sigma_t-D_\text{k})}{D_\text{in}}\right)
\begin{split}
R_A(&m_t, a_t,\sigma)=
R_A(m_t,a_t)\cdot 
\begin{cases} 
1 & \text{if $\sigma > D_\text{in}$}\\
\frac{\sigma}{D_\text{in}} & \text{else}\\
\end{cases}
\end{split}
\end{equation}
where $R_A(m_t, a_t)$ and $D_\text{in}$ are estimated by the learning modules. 
To save computation, we compute $D_\text{k}$, $D_\text{in}$ and $D_\text{out}$ as geometric distances on the occupancy grid.
To account for the dynamics of the agent navigating on a continuous space, we compute the ratio between geometric distance and timesteps by approximating the ratio between turning actions and moving forward steps as a constant.
In all experiments, we use an empirically computed ratio of 1.7, computed from offline navigation trials in training environments.

An illustration of the entire navigation system can be found in Fig.~\ref{fig:modules}.

%=============================================================================================================================================
\section{Learning-Based Modules}
\label{sec:learning}

\subsection{Estimating Properties of Unknown Space}
We estimate the values of $R_A$, $D_\text{in}$ and $D_\text{out}$ in Eq.~\eqref{eq:dp} using a neural network.
We generate the training values of $R_A$ for a frontier by computing the number of unobserved navigable cells on the current map beyond each frontier.
Given a partially observed map, we compute the neighboring unexplored area beyond each frontier via connected components and count the \deleted{number of}free cells within each region as $R_A$.

We compute the training values of $D_\text{in}$ and $D_\text{out}$ for a frontier on the skeleton of the occupancy map.
\added{
For skeleton computation, we use the approach described in~\cite{Zhang1984AFP} that iteratively removes boundary pixels while preserving topological structure.
}
We generate the skeleton on the free space of the ground-truth occupancy map.
We compute the subgraph of the skeleton for each frontier by taking nodes located in the frontier's neighbouring connected component.
We run a Travelling Salesman Problem solver on the subgraph to get the trajectory length of visiting every node once and returning to the frontier. 
We divide the entire travelling cost into two parts: $D_\text{in}$ is the travelling cost of visiting every node once and then stopping and $D_\text{out}$ is the cost of travelling back to the initial node from where $D_\text{in}$ finishes.

\subsection{Training Data Generation}

Since we learn the properties of the unknown space beyond each frontier, we require training data generated from a large number of environments.
We first build ground-truth occupancy and semantic maps for each environment.
We use an agent with a panoramic sensor running classical frontier-based exploration on these environments.
As the agent explores, revealing new portions of the environment, we extract the newly updated frontiers and compute their properties $R_A$, $D_\text{in}$ and $D_\text{out}$ as described in the previous section.
We also record the currently observed map $m_t$ and extract the egocentric observations $o_t$ corresponding to each frontier from the panorama.
We collect data tuple $(m_t, o_t, R_A, D_\text{in}, D_\text{out})$ for training the learning modules.

\subsection{Learning Module Architecture}
We build two learning models given different input representations, as shown by Fig.~\ref{fig:modules}.
We use a UNet~\cite{ronneberger2015u} architecture that inputs the current observed map $m_t$ and outputs a same-sized map with three channels.
Each channel of the output map represents the predictions of $R_A$, $D_\text{in}$ and $D_\text{out}$ at each location.
The loss is computed only on the frontier locations as other areas have no values.
\added{During testing stage, we average the predictions at the frontier pixels to get the estimated properties.}
We train another ResNet18 model that takes in the egocentric observations $o_t$ (a depth image and a semantic segmentation image) and outputs three values as predictions of $R_A$, $D_\text{in}$ and $D_\text{out}$.
We train these two models through supervised learning with L1-loss.

%=============================================================================================================================================
\section{Experiments}
\label{sec:exp}

\noindent \textbf{Datasets + Simulated Environments} We use the Habitat~\cite{savva2019habitat} simulator with the Matterport (MP3D)~\cite{Chang2017Matterport3DLF} dataset for our experiments. 
MP3D consists of 90 scenes that are 3D reconstructions of real-world environments, and semantic segmentations (40 semantic categories) of egocentric views are available.
We split the MP3D scenes into train:val:test following the standard 61:11:18 ratio to train the perception models and evaluate the explorers.
We split the 18 test scenes into small (size below $200m^2$), medium (size between $200m^2$ and $500m^2$) and large (size over $500m^2$) when reporting the results to better understand the advantages of our approach as the environment becomes larger.
We evaluate different exploration approaches using the standard 1000 testing episodes generated for the Habitat PointGoal Challenge~\cite{savva2019habitat}.
For our experiments, noise-free poses and ground-truth egocentric semantic segmentations are provided by the simulator.

\noindent \textbf{Task Setup} The robot observation space consists of RGB, depth and semantic segmentation images.
As our approach's performance largely depends on the FoV of the sensors, we evaluate our methods with two types of sensors:
(i) a panoramic $360^{\circ}$ FoV sensor with observations of $256 \times 1024$ resolution.
(ii) an egocentric $90^{\circ}$ FoV sensor with observations of $256 \times 256$ resolution.
The action space contains four short-term actions: \texttt{MOVE\_FORWARD} by 25cm, \texttt{TURN\_LEFT} and \texttt{TURN\_RIGHT} by $30^{\circ}$ and \texttt{STOP}.

%\noindent \textbf{Metric} 
We evaluate the explorers with the \textit{Coverage (Cov)} metric: the percentage of explored space over the entire environment within the allowed 500 steps.

\noindent \textbf{Implementation Details} 
We use the same mapping, frontier computation and local navigation implementations for all the baseline methods and our methods.
For training the learning modules, we pre-compute 220,000 train/ 1,000 val data tuples.
We train the model using PyTorch for six epochs and use the Adam optimizer with a learning rate of 0.001 decayed by a factor of 10 after every two epochs.
Training is done with two A100 GPUs within 12 hours.
\added{A single step in a navigation episode requires 0.9s on average including mapping, Bellman Equation computation, and local navigation. 
The timing was performed on a server using Xeon CPU @ 2.30GHz and an A100 GPU.}

\begin{figure}[t]
\centering
\includegraphics[width=0.48\textwidth]{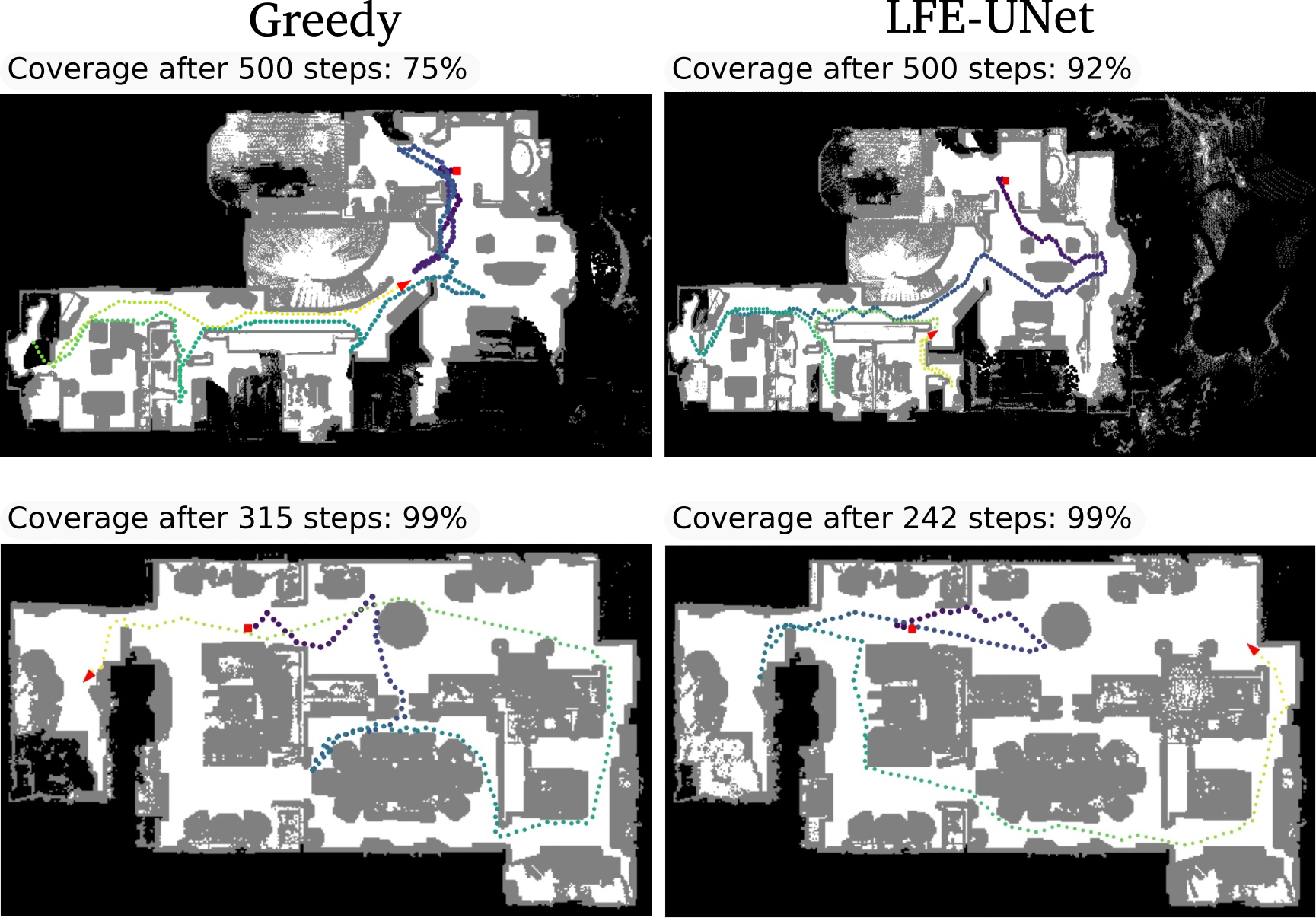}
\vspace{-2em}
\caption{
\textbf{A comparison between the coverage of our LFE-UNet explorer and the PONI-Greedy baseline on the test scenes.}
On a large scene (top row), LFE-UNet outperforms the Greedy baseline by $17\%$ on coverage after 500 steps.
On a small scene (bottom row), LFE-UNet finishes exploring the entire environment with 73 fewer steps than the Greedy baseline.
}\vspace{-1em}
\label{fig:example_traj}
\end{figure}

\subsection{Exploration with a Panoramic $360^{\circ}$ FoV Sensor}
\label{sec:exp_360}
We evaluate the explorers' performance when the robot's observation is a panorama.
We compare our method with the following three baselines:\\
\noindent\textbf{Frontier-Based Exploration (FBE-Near)}~\cite{yamauchi1997frontier}: A non-learned greedy algorithm. The agent always navigates to the nearest frontier.\\
\noindent\textbf{Active Neural SLAM (ANS+FBE)}~\cite{chaplot2020learning}: A learned exploration model trained with model-free Deep RL using coverage as rewards.
We use this as a comparison between model-free and model-based planning.
It has a global policy module for selecting long-term goals given the currently observed occupancy map and agent-visited location history as input.
We use our self-built map as input to their global policy with their trained weights to get the long-term goals 
and select the frontier nearest to the long-term goal for robustness consideration and navigate towards it.
We follow the original paper's setup by estimating a long-term goal every 25 steps or after the robot reaches the frontier.\\
\noindent\textbf{PONI-Greedy}~\cite{ramakrishnan2022poni}: PONI is a recent approach to frontier-based exploration that navigates greedily to the frontier with the maximum spatial and semantic potentials.
We adapt our implementation for this baseline by only using the learned $R_A$ of UNet, and the agent always chooses the frontier with the maximum estimated area.

\begin{table}[t]
\caption{coverage($\%$) of exploration methods with a panoramic $360^{\circ}$ FOV Sensor (higher is better)}
\label{tab:explore_360}
\centering
\begin{tabular}{l|c c c c}
Dataset(Scenes) & \multicolumn{1}{c}{Small} & \multicolumn{1}{c}{Medium}  & \multicolumn{1}{c}{Large} & \multicolumn{1}{c}{Overall}\\ 
\hline
%Method & Cov ($\%$)$\uparrow$  & Cov ($\%$)$\uparrow$  & Cov ($\%$)$\uparrow$  & Cov ($\%$)$\uparrow$  \\ 
%\hline
FBE-Near~\cite{yamauchi1997frontier} & 92.7 & 74.8  & 61.0  & 76.8  \\

ANS+FBE~\cite{chaplot2020learning} & 92.7  & 63.4  & 60.2  & 72.3  \\ 

PONI-Greedy~\cite{ramakrishnan2022poni} & 95.8  & 72.3  & 56.7  & 74.4  \\ 

LFE-UNet (Ours)  & \textbf{97.4}  & \textbf{75.1}  & \textbf{70.9}  & \textbf{80.7} \\ 

LFE-View (Ours) & 92.7  & 67.3  & 60.9  & 73.3  \\
\hline
Ablations: \\
LFE-($R^o,D^o$) & 95.2   & 84.0  & 79.5  & 86.2  \\ 
Greedy-($R^o$) & 90.1   & 57.3  & 35.1  & 61.7  \\ 
\end{tabular}\vspace{-1em}
\end{table}

We show these results in Table~\ref{tab:explore_360}.
Our LFE-UNet model outperforms all the other approaches.
We show example trajectories of LFE-UNet and PONI-Greedy in Fig.~\ref{fig:example_traj}.
LFE-View performs slightly worse, indicating that maps are more effective input representations to learn the frontier properties than the views.

\subsection{Exploration with an Egocentric $90^{\circ}$ FoV Sensor}
We evaluate the same set of approaches as in Sec.~\ref{sec:exp_360} with a $90^{\circ}$ FoV sensor, as this is a more common setup in recent exploration papers~\cite{Ramakrishnan2020AnEO, Georgakis2022UncertaintydrivenPF}.
Even though direct observations of many frontiers are missing, the learning-based approaches can still estimate the properties of the frontiers from the maintained map.
Owing to the limited field of view for these trials, we do not evaluate with the ResNet18 (the LFE-View planner), which requires a panoramic view that can see all frontiers.
Moreover, many recent exploration approaches~\cite{Georgakis2022UncertaintydrivenPF, ramakrishnan2020occupancy} are more focused on building an accurate map from a learned map module, something we compute directly from RGB-D and semantic information; as such, direct comparison with these approaches was not made.

The results are reported in Table~\ref{tab:explore_90}.
While the relative advantage of our method is reduced with sensors of limited FoV, our method still achieves the best overall coverage compared to the other methods.

\subsection{Ablation Study}
We study the impact of our learning modules, which estimate $R_A$, $D_\text{in}$, and $D_\text{out}$, by endowing the learning-augmented models with an oracle, which produces the true values of the area $R^o_A$ and distances $D^o = \{D^o_\text{in}, D^o_\text{out}\}$ beyond the frontiers.
These models are viewed as an upper performance bound for our approach.
We use the $360^{\circ}$ FoV sensor in this experiment.

The results are included in the last two rows of Table~\ref{tab:explore_360} and show that overall our oracle-informed model LFE-($R^o$, $D^o$) outperforms the other approaches and an oracle-informed greedy planner Greedy-($R^o$).
This empirically shows that there is still room for improvement in our learning module.
The oracle model can explore the large scenes $18.5\%$ more efficiently than the FBE-Near approach.
This is useful in real-world applications when a robot is sent to review the environment with a previously-built map.

The performance of the greedy approach drops with the known values since the true value of the area beyond multiple frontiers may exactly match, causing numerical instability during selection.
Thus the greedy heuristic oscillates between these frontiers, delaying exploration progress.
We found similar oscillations with the ANS+FBE~\cite{chaplot2020learning} approach, also trained with a greedy strategy. 
This oscillation problem is primarily reduced by using the learning modules because the prediction is imperfect, so two frontiers are unlikely to have the same predicted values.
Our approach avoids this unstable behavior through non-myopic planning and incorporating the time to travel to nearby frontiers.

\begin{table}[t]
\caption{coverage ($\%$) of exploration methods with a $90^{\circ}$ fov sensor}
\label{tab:explore_90}
\centering
\begin{tabular}{l | c c c c}

Dataset(Scenes) & Small & Medium  & Large & Overall\\ 
\hline
%Method & Cov ($\%$)$\uparrow$ & Cov ($\%$)$\uparrow$  & Cov ($\%$)$\uparrow$  & Cov ($\%$)$\uparrow$  \\ 
%\hline
FBE-Near~\cite{yamauchi1997frontier} & 84.0   & 60.3  & 51.3  & 65.2  \\ 

ANS+FBE~\cite{chaplot2020learning} & 93.0 & 47.9 & 44.4  & 61.1\\ 

PONI-Greedy~\cite{ramakrishnan2022poni} & 93.8  & \textbf{63.6}  & 47.5  & 68.8  \\ 

LFE-UNet (Ours)  & \textbf{96.9}  & 63.5  & \textbf{53.3}  & \textbf{70.9}  \\ 
%\hline

\end{tabular}\vspace{-1em}
\end{table}

\subsection{Learning Module Evaluation}
Here we study whether semantics helps with estimating the frontier properties.
We evaluate the performance of UNet by varying the input representations.
We tried two types of input: one with only an occupancy map versus having both an occupancy and a semantic map.
We trained another UNet model using the same training setup but only the occupancy map as input.
We compute the mean absolute error (MAE) of $R_A$, $D_\text{in}$ and $D_\text{out}$ altogether over all the frontiers in the testing examples.
With only the occupancy map as input, UNet achieves 156.8 MAE error while the MAE error is 140.0 with the semantic map.
This matches our intuition that having semantics in the perception system helps the learning module better estimate frontier properties of the indoor scenes.

%=============================================================================================================================================
\section{Conclusion}
We introduce a novel model-based approach for the time-limited exploration task in previously unseen environments.
Our approach augments classical frontier-based exploration with predictions about unseen space using learning.
Exploiting advances in visual sensing, semantic mapping and availability of large datasets of 3D models  
of real indoor scenes, we showed how to use a deep convolutional neural network
to learn two informative properties of each frontier: (i) the navigable area in the unknown space beyond a frontier and (ii) the timestep cost of exploring this unknown region and coming back.
We tried two input alternatives to the learning modules: (i) top-down-view occupancy and semantic map and (ii) egocentric depth and semantic segmentation views.
Our exploration experiments show that our method is $2.1\%$ more efficient in covering unknown areas than greedy strategies and $8.4\%$ more than myopic RL approaches.

%\section{Characterizing Explorer Behaviors}

\bibliographystyle{IEEEtran}
\bibliography{IEEEfull,main}

\begin{thebibliography}{10}
\providecommand{\url}[1]{#1}
\csname url@samestyle\endcsname
\providecommand{\newblock}{\relax}
\providecommand{\bibinfo}[2]{#2}
\providecommand{\BIBentrySTDinterwordspacing}{\spaceskip=0pt\relax}
\providecommand{\BIBentryALTinterwordstretchfactor}{4}
\providecommand{\BIBentryALTinterwordspacing}{\spaceskip=\fontdimen2\font plus
\BIBentryALTinterwordstretchfactor\fontdimen3\font minus
  \fontdimen4\font\relax}
\providecommand{\BIBforeignlanguage}[2]{{%
\expandafter\ifx\csname l@#1\endcsname\relax
\typeout{** WARNING: IEEEtran.bst: No hyphenation pattern has been}%
\typeout{** loaded for the language `#1'. Using the pattern for}%
\typeout{** the default language instead.}%
\else
\language=\csname l@#1\endcsname
\fi
#2}}
\providecommand{\BIBdecl}{\relax}
\BIBdecl

\bibitem{Ramakrishnan2020AnEO}
S.~K. Ramakrishnan, D.~Jayaraman, and K.~Grauman, ``An exploration of embodied
  visual exploration,'' \emph{International Journal of Computer Vision}, vol.
  129, pp. 1616 -- 1649, 2020.

\bibitem{Chaplot2020NeuralTS}
D.~S. Chaplot, R.~Salakhutdinov, A.~K. Gupta, and S.~Gupta, ``Neural
  topological {SLAM} for visual navigation,'' \emph{2020 IEEE/CVF Conference on
  Computer Vision and Pattern Recognition (CVPR)}, 2020.

\bibitem{Georgakis2022UncertaintydrivenPF}
G.~Georgakis, B.~Bucher, A.~Arapin, K.~Schmeckpeper, N.~Matni, and
  K.~Daniilidis, ``Uncertainty-driven planner for exploration and navigation,''
  \emph{2022 International Conference on Robotics and Automation (ICRA)}, pp.
  11\,295--11\,302, 2022.

\bibitem{ramakrishnan2022poni}
S.~K. Ramakrishnan, D.~S. Chaplot, Z.~Al-Halah, J.~Malik, and K.~Grauman,
  ``{PONI}: Potential functions for {ObjectGoal} navigation with
  interaction-free learning,'' in \emph{Proceedings of the IEEE/CVF Conference
  on Computer Vision and Pattern Recognition}, 2022.

\bibitem{bradley2021learning}
C.~Bradley, A.~Pacheck, G.~J. Stein, S.~Castro, H.~Kress-Gazit, and N.~Roy,
  ``Learning and planning for temporally extended tasks in unknown
  environments,'' in \emph{2021 IEEE International Conference on Robotics and
  Automation (ICRA)}.\hskip 1em plus 0.5em minus 0.4em\relax IEEE, 2021, pp.
  4830--4836.

\bibitem{chaplot2020learning}
D.~S. Chaplot, D.~Gandhi, S.~Gupta, A.~Gupta, and R.~Salakhutdinov, ``Learning
  to explore using active neural mapping,'' in \emph{International Conference
  on Learning Representations}, 2020.

\bibitem{chen2018learning}
T.~Chen, S.~Gupta, and A.~Gupta, ``Learning exploration policies for
  navigation,'' in \emph{International Conference on Learning Representations},
  2019.

\bibitem{Chang2017Matterport3DLF}
A.~X. Chang, A.~Dai, T.~A. Funkhouser, M.~Halber, M.~Nie{\ss}ner, M.~Savva,
  S.~Song, A.~Zeng, and Y.~Zhang, ``{Matterport3D}: Learning from {RGB-D} data
  in indoor environments,'' \emph{2017 International Conference on 3D Vision
  (3DV)}, pp. 667--676, 2017.

\bibitem{xia2018gibson}
F.~Xia, A.~R. Zamir, Z.~He, A.~Sax, J.~Malik, and S.~Savarese, ``Gibson {Env}:
  Real-world perception for embodied agents,'' in \emph{Proceedings of the IEEE
  Conference on Computer Vision and Pattern Recognition}, 2018, pp. 9068--9079.

\bibitem{yamauchi1997frontier}
B.~Yamauchi, ``A frontier-based approach for autonomous exploration,'' in
  \emph{Proceedings 1997 IEEE International Symposium on Computational
  Intelligence in Robotics and Automation CIRA'97.'Towards New Computational
  Principles for Robotics and Automation'}.\hskip 1em plus 0.5em minus
  0.4em\relax IEEE, 1997.

\bibitem{stein2018learning}
G.~J. Stein, C.~Bradley, and N.~Roy, ``Learning over subgoals for efficient
  navigation of structured, unknown environments,'' in \emph{Conference on
  Robot Learning}.\hskip 1em plus 0.5em minus 0.4em\relax PMLR, 2018, pp.
  213--222.

\bibitem{kaelbling1998planning}
L.~P. Kaelbling, M.~L. Littman, and A.~R. Cassandra, ``Planning and acting in
  partially observable stochastic domains,'' \emph{Artificial Intelligence},
  vol. 101, no. 1-2, pp. 99--134, 1998.

\bibitem{russell2010artificial}
S.~J. Russell, \emph{Artificial intelligence a modern approach}.\hskip 1em plus
  0.5em minus 0.4em\relax Pearson Education, Inc., 2010.

\bibitem{savva2019habitat}
M.~Savva, A.~Kadian, O.~Maksymets, Y.~Zhao, E.~Wijmans, B.~Jain, J.~Straub,
  J.~Liu, V.~Koltun, J.~Malik \emph{et~al.}, ``Habitat: A platform for embodied
  {AI} research,'' in \emph{Proceedings of the IEEE/CVF International
  Conference on Computer Vision}, 2019, pp. 9339--9347.

\bibitem{deitke2022retrospectives}
M.~Deitke, D.~Batra, Y.~Bisk, T.~Campari, A.~X. Chang, D.~S. Chaplot, C.~Chen,
  C.~P. D'Arpino, K.~Ehsani, A.~Farhadi \emph{et~al.}, ``Retrospectives on the
  embodied {AI} workshop,'' \emph{arXiv preprint arXiv:2210.06849}, 2022.

\bibitem{szot2021habitat}
A.~Szot, A.~Clegg, E.~Undersander, E.~Wijmans, Y.~Zhao, J.~Turner, N.~Maestre,
  M.~Mukadam, D.~Chaplot, O.~Maksymets, A.~Gokaslan, V.~Vondrus, S.~Dharur,
  F.~Meier, W.~Galuba, A.~Chang, Z.~Kira, V.~Koltun, J.~Malik, M.~Savva, and
  D.~Batra, ``Habitat 2.0: Training home assistants to rearrange their
  habitat,'' in \emph{Advances in Neural Information Processing Systems
  (NeurIPS)}, 2021.

\bibitem{habitat19iccv}
M.~Savva, A.~Kadian, O.~Maksymets, Y.~Zhao, E.~Wijmans, B.~Jain, J.~Straub,
  J.~Liu, V.~Koltun, J.~Malik, D.~Parikh, and D.~Batra, ``Habitat: {A}
  {P}latform for {E}mbodied {AI} {R}esearch,'' in \emph{Proceedings of the
  IEEE/CVF International Conference on Computer Vision}, 2019.

\bibitem{kolve2017ai2}
E.~Kolve, R.~Mottaghi, W.~Han, E.~VanderBilt, L.~Weihs, A.~Herrasti, D.~Gordon,
  Y.~Zhu, A.~Gupta, and A.~Farhadi, ``{AI2-THOR}: An interactive {3D}
  environment for visual {AI},'' \emph{arXiv preprint arXiv:1712.05474}, 2017.

\bibitem{ramakrishnan2021hm3d}
\BIBentryALTinterwordspacing
S.~K. Ramakrishnan, A.~Gokaslan, E.~Wijmans, O.~Maksymets, A.~Clegg, J.~M.
  Turner, E.~Undersander, W.~Galuba, A.~Westbury, A.~X. Chang, M.~Savva,
  Y.~Zhao, and D.~Batra, ``Habitat-{Matterport} 3{D} dataset ({HM}3d): 1000
  large-scale {3D} environments for embodied {AI},'' in \emph{Thirty-fifth
  Conference on Neural Information Processing Systems Datasets and Benchmarks
  Track}, 2021. [Online]. Available: \url{https://arxiv.org/abs/2109.08238}
\BIBentrySTDinterwordspacing

\bibitem{yadav2022habitat}
K.~Yadav, R.~Ramrakhya, S.~K. Ramakrishnan, T.~Gervet, J.~Turner, A.~Gokaslan,
  N.~Maestre, A.~X. Chang, D.~Batra, M.~Savva \emph{et~al.},
  ``Habitat-{Matterport} {3D} semantics dataset,'' \emph{arXiv preprint
  arXiv:2210.05633}, 2022.

\bibitem{wijmans2019dd}
E.~Wijmans, A.~Kadian, A.~Morcos, S.~Lee, I.~Essa, D.~Parikh, M.~Savva, and
  D.~Batra, ``{DD-PPO}: Learning near-perfect {PointGoal} navigators from 2.5
  billion frames,'' in \emph{International Conference on Learning
  Representations}, 2019.

\bibitem{Partsey2022IsMN}
R.~Partsey, E.~Wijmans, N.~Yokoyama, O.~Dobosevych, D.~Batra, and O.~Maksymets,
  ``Is mapping necessary for realistic {PointGoal} navigation?'' \emph{2022
  IEEE/CVF Conference on Computer Vision and Pattern Recognition (CVPR)}, pp.
  17\,211--17\,220, 2022.

\bibitem{li2022comparison}
\BIBentryALTinterwordspacing
Y.~Li, A.~Debnath, G.~J. Stein, and J.~Kosecka, ``Comparison of model free and
  model-based learning-informed planning for {PointGoal} navigation,'' in
  \emph{CoRL 2022 Workshop on Learning, Perception, and Abstraction for
  Long-Horizon Planning}, 2022. [Online]. Available:
  \url{https://openreview.net/forum?id=2s92OhjT4L}
\BIBentrySTDinterwordspacing

\bibitem{pal2021learning}
A.~Pal, Y.~Qiu, and H.~Christensen, ``Learning hierarchical relationships for
  object-goal navigation,'' in \emph{Conference on Robot Learning}.\hskip 1em
  plus 0.5em minus 0.4em\relax PMLR, 2021, pp. 517--528.

\bibitem{chaplot2020object}
D.~S. Chaplot, D.~P. Gandhi, A.~Gupta, and R.~R. Salakhutdinov, ``Object goal
  navigation using goal-oriented semantic exploration,'' \emph{Advances in
  Neural Information Processing Systems}, vol.~33, 2020.

\bibitem{savinov2018semi}
N.~Savinov, A.~Dosovitskiy, and V.~Koltun, ``Semi-parametric topological memory
  for navigation,'' \emph{arXiv preprint arXiv:1803.00653}, 2018.

\bibitem{Hahn2021NoRN}
M.~Hahn, D.~S. Chaplot, and S.~Tulsiani, ``No {RL}, no simulation: Learning to
  navigate without navigating,'' in \emph{Neural Information Processing
  Systems}, 2021.

\bibitem{li2020learning}
Y.~Li and J.~Ko{\v{s}}ecka, ``Learning view and target invariant visual
  servoing for navigation,'' in \emph{2020 IEEE International Conference on
  Robotics and Automation (ICRA)}.\hskip 1em plus 0.5em minus 0.4em\relax IEEE,
  2020, pp. 658--664.

\bibitem{Krantz2021WaypointMF}
J.~Krantz, A.~Gokaslan, D.~Batra, S.~Lee, and O.~Maksymets, ``Waypoint models
  for instruction-guided navigation in continuous environments,'' \emph{2021
  IEEE/CVF International Conference on Computer Vision (ICCV)}, pp.
  15\,142--15\,151, 2021.

\bibitem{krantz2022iterative}
J.~Krantz, S.~Banerjee, W.~Zhu, J.~Corso, P.~Anderson, S.~Lee, and J.~Thomason,
  ``Iterative vision-and-language navigation,'' \emph{arXiv preprint
  arXiv:2210.03087}, 2022.

\bibitem{batra2020rearrangement}
D.~Batra, A.~X. Chang, S.~Chernova, A.~J. Davison, J.~Deng, V.~Koltun,
  S.~Levine, J.~Malik, I.~Mordatch, R.~Mottaghi \emph{et~al.}, ``Rearrangement:
  A challenge for embodied {AI},'' \emph{arXiv preprint arXiv:2011.01975},
  2020.

\bibitem{Sarch2022TIDEETU}
G.~Sarch, Z.~Fang, A.~W. Harley, P.~Schydlo, M.~J. Tarr, S.~Gupta, and
  K.~Fragkiadaki, ``{TIDEE}: Tidying up novel rooms using visuo-semantic
  commonsense priors,'' in \emph{European Conference on Computer Vision}, 2022.

\bibitem{gu2022multi}
J.~Gu, D.~S. Chaplot, H.~Su, and J.~Malik, ``Multi-skill mobile manipulation
  for object rearrangement,'' in \emph{Deep Reinforcement Learning Workshop
  NeurIPS 2022}, 2022.

\bibitem{fang2019scene}
K.~Fang, A.~Toshev, L.~Fei-Fei, and S.~Savarese, ``Scene memory transformer for
  embodied agents in long-horizon tasks,'' in \emph{Proceedings of the IEEE/CVF
  Conference on Computer Vision and Pattern Recognition}, 2019, pp. 538--547.

\bibitem{ramakrishnan2020occupancy}
S.~K. Ramakrishnan, Z.~Al-Halah, and K.~Grauman, ``Occupancy anticipation for
  efficient exploration and navigation,'' in \emph{European Conference on
  Computer Vision}.\hskip 1em plus 0.5em minus 0.4em\relax Springer, 2020, pp.
  400--418.

\bibitem{chen2020learning}
C.~Chen, S.~Majumder, Z.~Al-Halah, R.~Gao, S.~K. Ramakrishnan, and K.~Grauman,
  ``Learning to set waypoints for audio-visual navigation,'' in
  \emph{International Conference on Learning Representations}, 2020.

\bibitem{Pineau-2002-8519}
J.~Pineau and S.~Thrun, ``An integrated approach to hierarchy and abstraction
  for {POMDPs},'' Carnegie Mellon University, Pittsburgh, PA, Tech. Rep.
  CMU-RI-TR-02-21, August 2002.

\bibitem{sethian1996fast}
J.~A. Sethian, ``A fast marching level set method for monotonically advancing
  fronts.'' \emph{Proceedings of the National Academy of Sciences}, vol.~93,
  no.~4, pp. 1591--1595, 1996.

\bibitem{Zhang1984AFP}
T.~Y. Zhang and C.~Y. Suen, ``A fast parallel algorithm for thinning digital
  patterns,'' \emph{Commun. ACM}, vol.~27, pp. 236--239, 1984.

\bibitem{ronneberger2015u}
O.~Ronneberger, P.~Fischer, and T.~Brox, ``{U-Net}: Convolutional networks for
  biomedical image segmentation,'' in \emph{International Conference on Medical
  image computing and computer-assisted intervention}.\hskip 1em plus 0.5em
  minus 0.4em\relax Springer, 2015, pp. 234--241.

\end{thebibliography}
%\bibliography{main}

\end{document}